\documentclass{article}
\usepackage{ijcai19}
\usepackage[utf8]{inputenc}
\usepackage{CJK}
\input zhwinfonts

\usepackage{times}
\usepackage{soul}
\usepackage{url}
\usepackage[hidelinks]{hyperref}
\usepackage[small]{caption}
\usepackage{graphicx}
\usepackage{booktabs}
\usepackage[symbol]{footmisc}
\renewcommand{\thefootnote}{\fnsymbol{footnote}}

\usepackage{algorithm,algorithmic}
\usepackage{bm,array,mathtools,cases,thmtools,thm-restate}
\usepackage{amsthm,amssymb,amsmath}
\usepackage{longtable,rotating}

\title{Business Taxonomy Construction\\ Using Concept-Level Hierarchical Clustering}

\author{
Haodong Bai,$^\dag$ Frank Z. Xing,$^\ddag$ Erik Cambria,$^\ddag$ Win-Bin Huang$^{\dag}$\footnote{Corresponding author: Win-Bin Huang}\\
\affiliations
$^\dag$Department of Information Management, Peking University\\
$^\ddag$School of Computer Science and Engineering, Nanyang Technological University
\emails
\{hbai,huangwb\}@pku.edu.cn, \ \ \{zxing001,cambria\}@ntu.edu.sg
}

\begin{document}

\maketitle

\begin{abstract}
Business taxonomies are indispensable tools for investors to do equity research and make professional decisions. However, to identify the structure of industry sectors in an emerging market is challenging for two reasons. First, existing taxonomies are designed for mature markets, which may not be the appropriate classification for small companies with innovative business models. Second, emerging markets are fast-developing, thus the static business taxonomies cannot promptly reflect the new features. In this article, we propose a new method to construct business taxonomies automatically from the content of corporate annual reports. Extracted concepts are hierarchically clustered using greedy affinity propagation. Our method requires less supervision and is able to discover new terms. Experiments and evaluation on the Chinese National Equities Exchange and Quotations (NEEQ) market show several advantages of the business taxonomy we build. Our results provide an effective tool for understanding and investing in the new growth companies.
\end{abstract}

\renewcommand{\thefootnote}{\arabic{footnote}}
\section{Introduction} \label{intro}
Business taxonomies are important knowledge management tools for investment activities. When comparing different equity assets on the financial markets, investors tend to classify companies according to their main business sectors, market performances, and the products they manufacture. To discover companies with great potentials to grow across different industries, only those in the same industry sector will adopt similar criteria for downstream analysis, such as financial statement analysis, profit prediction, price-earnings valuation and more~\cite{92-AWA-F}. To this end, accurate classification of companies is crucial to successful investments. Consequently, governments and financial authorities, as well as big companies, have developed a large number of different business taxonomies, which are usually widely applicable, coarsely-grained and almost static. However, these features are not appropriate for small and startup companies. These companies are often fast-growing, dynamically changing their business and focusing on a specific business. Therefore, traditional business taxonomies cannot reflect the whole landscape and emerging business. Beside the traditional business taxonomies, Chinese stock markets have yet another knowledge management tool called ``concept stock (\begin{CJK}{UTF8}{gbsn}概念股\end{CJK})". However, the concept labels are summarized by research teams and media, which means that they have already attracted much attention and over-represent blue chip stocks. Moreover, the concept labels are neither systematic nor hierarchical. One such influential label set is Tonghuashun's ``concept boards"~\footnote{http://q.10jqka.com.cn/gn/}. For small and startup companies, the current situation is that the valuation of such companies has to rely on concept labels transferred from the main domestic ``A" shares markets, which do not appropriately describe small companies. The companies listed at the Chinese National Equities Exchange and Quotations (NEEQ)~\footnote{The NEEQ is an over-the-counter (OTC) system for trading the shares of a public limited company that is not listed on either the Shenzhen or Shanghai stock exchanges, thus nicknamed ``The New Third Board (\begin{CJK}{UTF8}{gbsn}新三板\end{CJK})".} are typical examples. Compared to those ``A" share companies, the NEEQ listed companies rely even heavier on the inappropriate concept labels because there are no widely agreed market capitalization or enterprise multiple to them.

For the above-mentioned reasons, there is an urgent need for a more flexible business taxonomy to help with the investment decisions for small and new companies. The taxonomy can form benchmarks for thousands of different companies with innovative business models. Compared to the concept labels, a business taxonomy is not only helpful for investigating a specific company, but also beneficial to understand the relations between companies. There is already a large amount of studies on automatic taxonomy construction (ATC) for applications such as web search~\cite{atc-keywords}, question answering and refinement~\cite{refine}, advertising and recommendation systems, and knowledge organization~\cite{taxo-gen}. However, few of them concerns business taxonomy construction. On the other hand, studies that leverage natural language processing (NLP) or text mining to support investment either improve the current existing taxonomy~\cite{hoberg-16} or express the industry structure using other mathematical tools~\cite{gvs}. Unlike previous research, we propose a new method in this article that constructs a business taxonomy from scratch. The method extracts concept-level terms from the corporate annual reports, and computes the similarities between different terms. Based on the similarity matrix, the method recursively cluster terms into different strata. 

Our \emph{contributions} are tri-fold:
\begin{enumerate}
\item To the best of our knowledge, we pioneer the use of automatic taxonomy construction for the \emph{business classification and investment purposes}. Using concept-level terms instead of keywords, the method needs a low level of supervision because we leverage linguistic knowledge and a statistical model to extract and compare terms. No seed terms or their relations are required.
\item We use positive and unlabeled learning (PU learning) to further mitigate the labor to tag indexing terms. The method thus shows its capability to identify fine-grained concepts and discover new terms from natural language.
\item We make the NEEQ annual reports dataset publicly available~\footnote{The dataset is downloadable from the following link: http://github.com/SenticNet/neeq-annual-reports/.}, such that researchers could benchmark their taxonomy construction methods on it or follow up with other text mining tasks. 
\end{enumerate}
The remainder of this article is organized as follows: Section~\ref{rwk} elaborates related work from two thread of literature: the business classification systems and studies on automatic taxonomy construction; Section~\ref{mthd} provides an overview of the framework and introduce details of the algorithm; Section~\ref{exp} presents experimental results; Section~\ref{eval-dis} evaluate the constructed taxonomy for the NEEQ market and carries out case studies; Finally, Section~\ref{cln} concludes the study with future directions. 

\section{Related Work} \label{rwk}
\subsection{Business Classification Systems}
Business classification systems, or industry classification schemes, are fundamental tools for market research. According to a recent review~\cite{phior}, companies are grouped and organized into categories by their similar manufacturing process, final products, and the target markets. Investors make use of the business classification systems for purposes such as benchmarking with flagship companies, discovering potential competitors, evaluating sales performances, and composing industry index. Mainstream business classification systems can be assorted into three classes depending on their developers and purposes: governmental statistical agencies develop the system for measuring economic activities, business information vendors develop the system for guiding investors, and academic researchers study the use of such system for accounting and finance. The most widely used examples are from business information producers, such as the Global Industry Classification Standard (GICS) and the Thomson Reuters Business Classification (TRBC), because they are integrated into the popular commercial databases. Early research~\cite{bhojraj} also supports that the GICS accurately classifies the market. For this reason, some business classification systems used on the Chinese financial markets are adapted from GICS, such as the SWS classification standard~\footnote{http://www.swsindex.com/pdf/swhylfsm.pdf/,\\ \indent Accessed on 2019-04-03.} and the official NEEQ classification guide~\footnote{http://www.neeq.com.cn/fenglei/hyfl.html/,\\ \indent Accessed on 2019-04-03.}. However, many problems have been found when using these systems on the NEEQ market. First, designed using a top-down approach, these systems have unbalanced numbers of companies in the end-level of classes. To fit in a pre-defined structure, many classes contain companies with different businesses. Second, small companies are still at the early stage of exploring their business strategies. Therefore, it is common that one company's business can span several domains in the system, while it can only be classified in a unique class. This causes the company's absence in other classes. Last yet importantly, frequent revision of such systems is costly and would confuse investors. 

Literature on using NLP and text mining for financial forecasting and investment activities is growing~\cite{17-XF-T}. Specific to business classification, Hoberg and Phillips built two systems using the 10-K corpus. The first one discovers competition relations between companies according to how similar are their product descriptions and constructs a company network~\cite{hoberg-10}. The second one first cluster companies with the text description of company products, then map the traditional business classification scheme to the newly constructed one~\cite{hoberg-16}. Both studies focused on improving the existing classification systems. Consequently, the details of a company's business model are not revealed and classification results are still rather coarse. Taxonomies with more detailed information, for example on products~\cite{e-commerce}, are not catered for the purpose of industry partition. In this research, we break the stereotype and take a fully data-driven approach for building the classification system based on the textual description of companies. The business-related concepts and terms are thus more detailed and information-rich.

\subsection{Automatic Taxonomy Construction}
A taxonomy is defined as a semantic hierarchy that organizes concepts by is-a relations~\cite{taxosur}. Since is-a relations are the most important relations in human cognitive structures, taxonomy construction from natural language is fundamental for ontology learning tasks. In common cases, ATC follows a pipeline of is-a relation extraction from natural language and induction of the taxonomy structure. 

Relation extraction can be either pattern-based or statistical. One of the pioneer pattern-based research by Hearst~\cite{hearst} proposed to use hand-crafted lexical patterns like ``A is a B" and  ``A such as B" to discover is-a relations. More syntactic patterns are proposed by following research~\cite{navigli,syntac-context}, for example, ``A, including B", ``A is a type/kind of B" etc. The performance can be improved by boosting over multiple such rules~\cite{boosting}. Pattern-based methods feature high precision but poor recall. This is because the exact match of such patterns has a low coverage over the relations contained in the corpus. This problem is more severe in our research because business descriptions usually do not contain explanatory clauses as above-mentioned in the linguistic patterns. Statistical model exams the relation between any two terms, i.e., first extract all the candidate terms, and build a model to predict what is the relation type or whether there exists an ``is-a" relation between two terms. The term extraction step can be achieved with either supervised or unsupervised machine learning algorithms. In the former case, more label of true terms will be required and in the latter, only minimum effort is taken to threshold terms using TF-IDF, topic modeling (LDA)~\cite{tm-taxo}, or TextRank model. For the relation predictive model, unsupervised methods leverage information such as co-occurrence frequency analysis, term subsumption~\cite{subsump-hc}, cosine similarity based on bag-of-words, and word embedding similarities~\cite{hwe} to discover taxonomic relations~\cite{taxosur}. Supervised methods require inductive reasoning over a set of known relations, which is more precise but rely heavily on the corpus as well as the seed relations~\cite{taxo-gen}. In some cases, supervised methods have very poor recall. Obviously, there is a trade-off between precision and recall.   

Induction of the taxonomy refers to the process of growing a graph-like structure based on the set of relations extracted from the previous step. The optimal taxonomy desires some features, such as no redundant edges and no loop of conceptual terms~\cite{syntac-context}. The most important objective is the correctness of hypernym-hyponym relations: comparable terms should belong to the same level. Practically speaking, the business taxonomy should provide the necessary knowledge and business insights pertinent to the investment activities. To enable these, current approaches employ either clustering or algorithms that induct tree structure from a graph. Clustering methods assume that agglomerated terms share the same hypernym. By recursively choosing a representative term, hierarchical clustering can generate a layered tree structure~\cite{subsump-hc,dss}. On the other hand, the term relations can be organized as a directed graph. Then the task becomes mining and pruning a tree structure out of the graph~\cite{choi}. In this research, we use a weakly supervised statistical method for relation extraction and greedy hierarchical affinity propagation (GHAP) to construct a new taxonomy, and relate companies to the leaf descendant layer.

\section{Methodology} \label{mthd}
\begin{figure*}[t]
\centering
\includegraphics[width=0.95\textwidth]{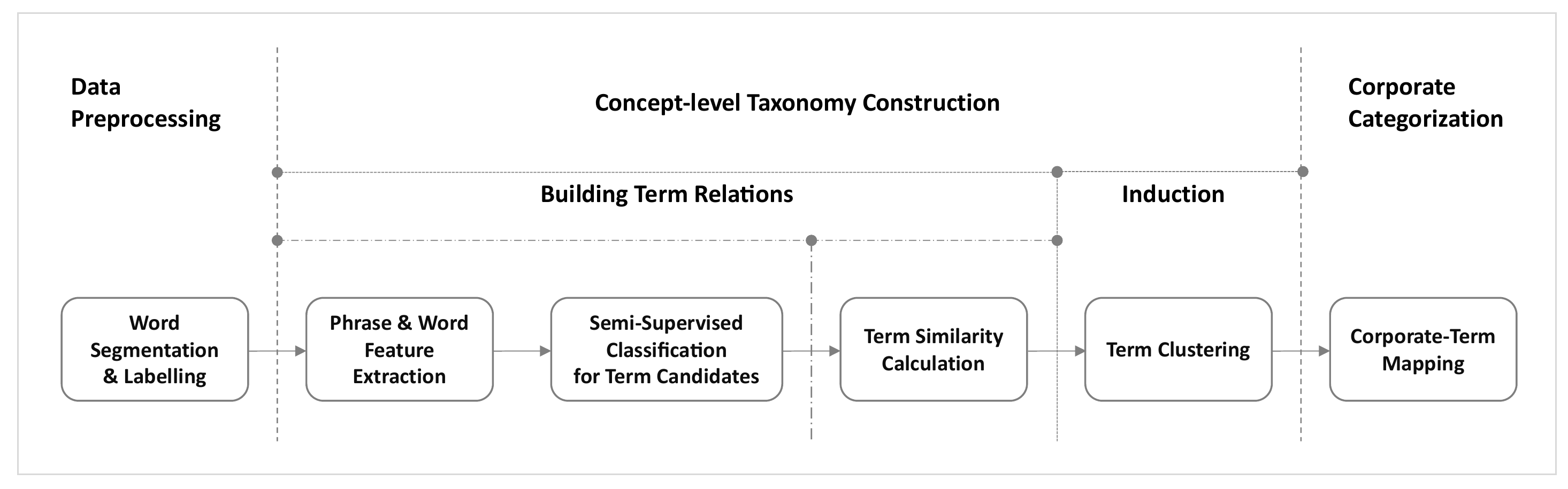}
\caption{An overview of the proposed method, showing key techniques used in each module.} \label{overviewf} 
\end{figure*}

Our method can be divided into three phases: data preprocessing, concept-level taxonomy construction, and corporate categorization and labeling with the established taxonomy. Figure~\ref{overviewf} provides an overview of the proposed method. Because the corpus we use is in Chinese, the data preprocessing phase consists of word segmentation and part-of-speech (POS) labeling of each Chinese word. We use the LTP-Cloud tools developed by HIT~\footnote{http://www.ltp-cloud.com/} to complete this phase. The taxonomy construction phase utilizes a semi-supervised learning classifier~\cite{pulearning} to reduce the amount of labor for tagging terms. After filtering out the concept term candidates, we obtain the final terms from the classifier. The similarity calculation is based on the idea of co-occurrence analysis from information science. Then GHAP takes the similarity matrix as an input to build a multi-layered structure of terms. The corporate categorization phase maps all the companies that contain the descendant-level terms to the taxonomy.

\begin{table}[t]
\centering
\caption{Concept-level features used to train a term extractor.} \label{concept-features}
\resizebox{0.48\textwidth}{!}{
\begin{tabular}{lp{6cm}}
\toprule
Name of features & Computing methods \\ 
\midrule
Concept mutual information & $MI(t)=\sum_{i, j} p(i, j) \times \log[p(i, j)/(p(i)p(j))]$.\\
Right-side entropy & $RE(t)=\sum_{i} p(t, i|t)\times \log(p(t, i|t))$.\\
Left-side entropy & $LE(t)=\sum_{i} p(i, t|t)\times \log(p(i, t|t))$.\\
Concept TF & The overall term frequency in all the documents.\\
Concept IDF & The overall inverse document frequency in all the documents.\\
Followed-by word & Binary feature of whether the concept is followed by ``industry (\begin{CJK}{UTF8}{gbsn}行业\end{CJK})" or ``business scope (\begin{CJK}{UTF8}{gbsn}业务\end{CJK})".\\
Following word & Binary feature of whether the concept is following ``running (\begin{CJK}{UTF8}{gbsn}从事\end{CJK})".\\
Industry TF & The concept frequency distribution in all the industry classes.\\
Industry IDF & The inverse document frequency distribution in all the industry classes.\\
Industry concept entropy & $IndE(t)=-\sum_{i} (TF_{t, i}/TF_{t})\times \log(TF_{t, i}/TF_{t})$.\\
\bottomrule
\end{tabular}}
\end{table}

\subsection{Concept Extraction and Term Similarity}
One of the fundamental challenges in NLP is to model the semantic compositionality within phrases and multi-word expressions. Previous research~\cite{14-CE-T} suggests considering concepts to be the atomic units of meaning, which leads to more powerful expressiveness and more accurate results in downstream applications. Unlike ATC study which uses keywords~\cite{atc-keywords}, we consider concept-level terms in our business taxonomy.

We observe that two types of templates together cover most of the concepts in the business domain, i.e. noun phrases and attributive phrases. For the first type, we mainly consider the noun-type POS tags in the ``863 Chinese POS set". Additionally, we include Chinese numerals~\footnote{Numerals appear in noun phrases such as ``Third-party payment (\begin{CJK}{UTF8}{gbsn}第三方支付\end{CJK})".} and verbs, which are not morphologically identifiable to ensure a high recall. For the second type, we simultaneously consider the dependency parsing result. Those phrases that only contains dependency relation ``ATT" (the attributive relation type in Chinese grammar) are selected to be concept term candidates.

The term candidates are represented with a concatenation of concept-level features as listed in Table~\ref{concept-features} and similar word-level features. The features are designed to include both statistical and industry-related information based on the official NEEQ classification guide, because the distribution of term frequencies in texts of different industries is a crucial fact to the discriminative power of the term.

The semi-supervised classifier is built as a support vector machine (SVM) with probabilistic outputs under the framework of PU learning~\cite{pulearning}. PU learning is calibrated for real-world problems where labels of the negative cases are not accessible. Labels for positive cases are costly and hard to exhaust, so the majority of data remains unlabeled. Through the analysis of the empirical risk minimization problem of SVM, it is proved that PU learning is equivalent to a cost-sensitive classification where the cost ratio $c_1/c_x$ is a function of class prior $\pi$ and proportion of labeled sample $\eta$~\cite{pulearning}:
\begin{equation}
    c_1/c_x = \frac{2\pi (1-\eta)}{\eta}.
\end{equation}
We use the scikit-learn package to implement the cost-sensitive SVM with RBF kernel and estimate the probability parameters from the dataset. In experiments, we use the dual problem settings of PU learning, where only a small portion of negative cases are labeled. This is made possible be checking if the term candidate contains words from the stop-word list. We adapt a general stop-word list to the specific business domain by adding 106 domain-specific words to it. The added words include common words in the business domain such as ``corporate (\begin{CJK}{UTF8}{gbsn}集团\end{CJK})", ``company (\begin{CJK}{UTF8}{gbsn}公司\end{CJK})" and action words such as ``sales (\begin{CJK}{UTF8}{gbsn}销售\end{CJK})", ``profit (\begin{CJK}{UTF8}{gbsn}盈利\end{CJK})", ``leading (\begin{CJK}{UTF8}{gbsn}领先\end{CJK})", ``trend (\begin{CJK}{UTF8}{gbsn}趋势\end{CJK})" etc. After training with the negative labels, the classifier produces the real term set from the term candidates.

A term similarity is computed by integrating the comprising word-level similarities. To be more specific, we define the similarity of two words as the frequency of their co-occurrence divided by the harmonic mean of the frequencies of their occurrence in the documents respectively. That is
\begin{equation}
    s(w_1, w_2) = \frac{2\times dct(w_1 \cap w_2) \times dct (w_1) \times dct (w_2)}{dct (w_1) + dct (w_2)},
\end{equation}
where $dct(\cdot)$ denotes document counts. Then, we align corresponding words in two terms and use the average similarity of the best-match as the similarity between terms. Because this method is asymmetric, we define term similarity as the average over two directions:
\begin{align}
s(t_1 \rightarrow t_2) &= \frac{\sum_{i \in t_1} \beta_i \max_{j \in t_2} s(i, j)}{len(t_1)}\\
s(t_1 , t_2)   &= \frac{s(t_1 \rightarrow t_2) + s(t_2 \rightarrow t_1)}{2}
\end{align}
where $i$ is word in term $t_1$ and $j$ is word in term $t_2$; $len(t_1)$ denotes the length of $t_1$. The weight for word $i$ uses the TF-IDF information:
\begin{equation}
   \beta_i  = \log(ct(i)) \times \log(\frac{N}{dct(i)}).
\end{equation}
where $N$ is the total number of documents.

\subsection{Taxonomy Induction}
The term similarity matrix measures semantic relations between two given terms, where the target ``is-a" relation is one of such. In order to construct a taxonomy, we computer a matrix of relations from the term similarity matrix by clustering, which preserve the strong relations while prune the others. We leverage greedy hierarchical affinity propagation (GHAP)~\cite{ghap}, an exemplar-based clustering method to construct three layers of hypernym-hyponym relations. Compared to other clustering method, such as K-means, GMM or DBSCAN, GHAP has some advantages for taxonomy construction. \emph{First}, the GHAP centroids are prototypical data points, which is important for the hypernym-hyponym relations. \emph{Second}, GHAP does not need the number of clusters as a hyper-parameter input. \emph{Third}, the clustering result of GHAP is insensitive to the initialization states. It is also worth mentioning that GHAP usually converges faster than HAP, which has to optimize a global loss function. The method is based on the concept of ``message passing" between data points. For each layer, we iteratively compute a availability matrix $\mathbb{A}[\alpha_{ij}]_{n\times n}$ and a responsibility matrix $\mathbb{R}[\rho_{ij}]_{n\times n}$~\cite{fd07}, where
\begin{align}
&\alpha_{ii} = c_i + \sum_{k\neq i} \max(0, \rho_{ki})\\
&\alpha_{ij}^{i\neq j} = \min[0, c_j + \rho_{jj} + \sum_{k\notin \{i,j\}} \max(0, \rho_{ki})]\\
&\rho_{ij} = s_{ij} - \max_{k\neq j}(\alpha_{ik}+s_{ik}),
\end{align}
$i$ and $j$ are taxonomic terms; $c_j$ is the preference for choosing term $j$ as an exemplar; $n$ is the number of terms or exemplar terms in that layer. The binary exemplar vector is subsequently obtained as $\mathbf{e} = (\mathrm{diag}(\mathbb{A})+\mathrm{diag}(\mathbb{R})>0)$. Each descendant term in this taxonomy further corresponds to a set of companies running similar business. A major difference of this taxonomy from traditional business classification systems is that one company can be mapped to multiple terms. This assumption is rational because in real-world cases, companies can span their business across several industry sectors.

\section{Experiments and Evaluation} \label{exp}
\subsection{Data and Results}
We crawled 21,739 annual reports for 10,375 listed companies from the NEEQ. The releasing time of these reports spans three years from 2014 to 2017. The original reports are in PDF format with relatively fixed discourse structure. We parse the files and extract texts from the section named ``business model" using Tabula~\footnote{http://tabula.technology/}. After manually cleaning the missing cases, we finally obtained 20,040 business model descriptions, summing up to 46.2 MB of textual data. According to the annual report standards, the descriptions cover the industry information, product and service, type of clients, key resource, sales model and components of income. Most of the descriptions comprise 100 to 1000 Chinese characters.

We obtained 64,460 concept-level term candidates from the corpus and labeled 7,078 of them as non-terms using the domain stop-word list. The cost-sensitive SVM classifier output 2,744 terms, which are clustered into 33 hypernyms (see Table~\ref{statshyp}). Our investigation shows that each hypernym governs no more than 20 sub-concept and 230 sub-sub-concept. Given the fact that the average term similarity equals 0.15, most of the clusters exhibit high intra-class similarity. We also observed a strong correlation between the numbers of sub- and sub-sub- concepts, which indicates the whole taxonomy is well-balanced. 

\begin{figure*}[tb]
\centering
\includegraphics[width=\textwidth]{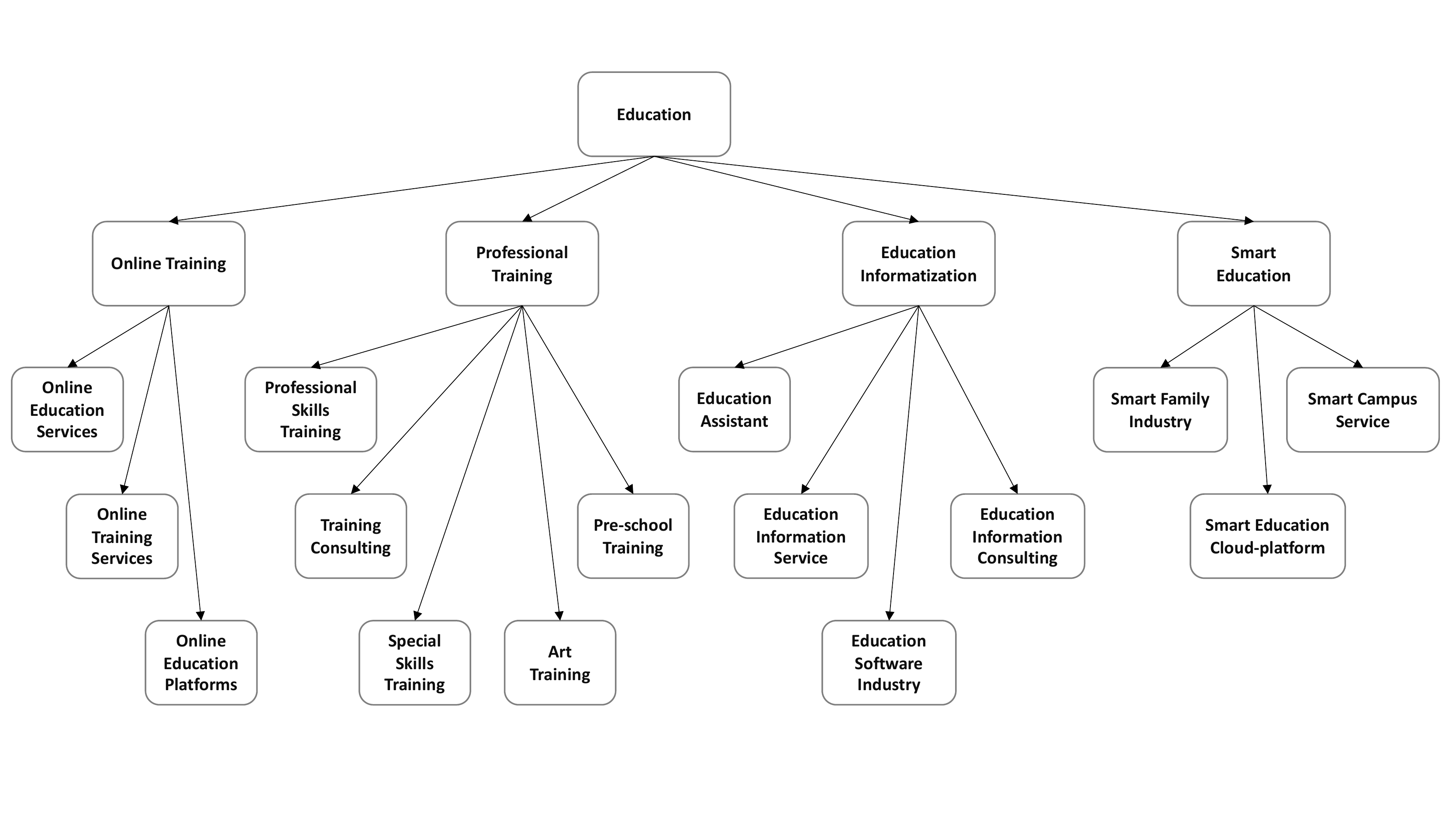}
\caption{Three level classification system for the education industry.} \label{edu} 
\end{figure*}

\begin{table}[t]
\centering
\caption{Statistics of the first level hypernyms.} \label{statshyp}
\resizebox{0.48\textwidth}{!}{
\begin{tabular}{p{6cm}p{1cm}p{1cm}p{1cm}p{1cm}}
\toprule
Hypernym & Intra-class similarity &No. of sub-concept & No. of sub-sub-concept & No. of companies\\
\midrule
Healthcare \begin{CJK}{UTF8}{gbsn}医疗诊断服务\end{CJK}& 0.40 &2&17&72\\
Education \begin{CJK}{UTF8}{gbsn}教育\end{CJK}& 0.37 &4&15&137\\
Lighting \begin{CJK}{UTF8}{gbsn}照明灯具\end{CJK}& 0.36 &4&34&147\\
Game \begin{CJK}{UTF8}{gbsn}游戏\end{CJK}& 0.34 &3&33&156\\
Transportation \& logistics \begin{CJK}{UTF8}{gbsn}物流运输\end{CJK}& 0.33 &3&22&206\\
Medical service \& equipment \begin{CJK}{UTF8}{gbsn}医疗器械制造与医疗服务\end{CJK}& 0.28 &5&22&353\\
Ironmongery \begin{CJK}{UTF8}{gbsn}金属零部件制造\end{CJK}& 0.27 &4&26&208\\
Software \& Hardware \begin{CJK}{UTF8}{gbsn}第三方软硬件\end{CJK}& 0.27 &4&51&525\\
Cement products \begin{CJK}{UTF8}{gbsn}金属混凝土产品\end{CJK}& 0.27 &3&9&34\\
Automobile \begin{CJK}{UTF8}{gbsn}汽车\end{CJK}& 0.25 &5&32&473\\
Electronics elements \begin{CJK}{UTF8}{gbsn}电子原件制造\end{CJK}& 0.24 &6&66&950\\
Telecoms \begin{CJK}{UTF8}{gbsn}通信及通信设备\end{CJK}& 0.24 &6&60&903\\
Building \begin{CJK}{UTF8}{gbsn}建筑工程\end{CJK}& 0.24 &7&59&433\\
Automation \& robotics \begin{CJK}{UTF8}{gbsn}自动化机器人\end{CJK}& 0.23 &3&21&169\\
Information system \& integration \begin{CJK}{UTF8}{gbsn}信息系统集成服务\end{CJK}& 0.23 &4&47&2416\\
Energy saving \begin{CJK}{UTF8}{gbsn}节能环保\end{CJK}& 0.23 &6&49&265\\
GIS service \begin{CJK}{UTF8}{gbsn}地理信息服务\end{CJK}& 0.22 &3&43&1601\\
IT infrastructure \& maintenance \begin{CJK}{UTF8}{gbsn}IT基础设施与运维\end{CJK}& 0.22 &4&32&252\\
Office appliance \begin{CJK}{UTF8}{gbsn}日常办公用品\end{CJK}& 0.22 &2&7&56\\
Digital media \begin{CJK}{UTF8}{gbsn}互联网数字媒体\end{CJK}& 0.22 &5&56&692\\
Clinical testing \begin{CJK}{UTF8}{gbsn}临床试验检测\end{CJK}& 0.21 &3&18&216\\
Smart houseware \begin{CJK}{UTF8}{gbsn}智能家居\end{CJK}& 0.21 &9&49&1086\\
Horticulture \begin{CJK}{UTF8}{gbsn}园林工程\end{CJK}& 0.20 &14&106&825\\
Mechanical equipment \begin{CJK}{UTF8}{gbsn}机械设备制造\end{CJK}& 0.20 &8&67&377\\
Chemicals \begin{CJK}{UTF8}{gbsn}化工产品\end{CJK}& 0.19 &6&35&274\\
Plastic products \begin{CJK}{UTF8}{gbsn}塑料制品\end{CJK}& 0.19 &12&59&395\\
Internet \& online ads \begin{CJK}{UTF8}{gbsn}互联网媒体广告\end{CJK}& 0.19 &13&106&1097\\
Solar battery \begin{CJK}{UTF8}{gbsn}太阳能电池\end{CJK}& 0.18 &19&188&1699\\
E-commerce platforms \begin{CJK}{UTF8}{gbsn}电商平台\end{CJK}& 0.17 &8&53&1568\\
Financial services \begin{CJK}{UTF8}{gbsn}金融服务\end{CJK}& 0.17 &10&78&2673\\
Outsourcing consulting \begin{CJK}{UTF8}{gbsn}工程咨询承包\end{CJK}& 0.17 &10&79&4154\\
Natural bio-extract \begin{CJK}{UTF8}{gbsn}天然植物提取物产品\end{CJK}& 0.16 &18&125&1194\\
Phone gadgets \begin{CJK}{UTF8}{gbsn}手机周边产品\end{CJK}& 0.16 &20&223&8876\\
\bottomrule
\end{tabular}}
\end{table}

To understand the branching structure within a hypernym, we showcase the structure of a relatively small ancestor class in the second row of Table~\ref{statshyp} --- ``Education" (see Figure~\ref{edu}). There are four sub-concepts attached to this class: online training, professional training, education informatization, and smart education. Each sub-concept also has several hyponyms. Due to limited space we can not include all the education industry companies. Instead, we compare some popular NEEQ classification label and terms produced by our method.

\subsection{Qualitative Evaluation and Discussion} \label{eval-dis}
We benchmark the validity of our constructed business taxonomy with the official NEEQ classification guide via human evaluation. Generally, the descendant classes in the traditional business classification system are coarse. For example, many companies in the online education or training scope are classified as ``Internet Software and Services", which is apparently wilder-ranging; similarly, some companies are labeled as ``General Customer Service", which provides less information than the concept of ``Online Training". In fact, ``Internet Software and Services" only reveals the means of conveying their product for online education companies. However, their customers, competitors, and market positioning are more comparable to traditional education companies, but are very different from internet software providers such as SAP or Tencent. In this sense, the traditional business classification system misleads investors by classifying companies with different business models together, providing inaccurate peers for pricing and research. In contrast, our method provides fine-grained concept-level terms. The mapping of companies are more balanced: each descendant term governs around ten companies in Table~\ref{statshyp}. 

Another important aim of investment analysis is to discover new concepts and market trends. The new concepts often reflect how the industries will re-organize and develop in the future. However, the low frequency of update for traditional business classification systems tends to hide new business concepts. It is also challenging to find the appropriate position for new concepts. We notice that the business owners tend to advertise the hotspot concepts in their self-descriptions. Because our method is aware of the content of corporate annual reports, new concepts can be captured during taxonomy construction. For example, ``online training" and ``education informatization" are trendy concepts in the scope of education. Pre-school training is also increasingly popular in China, probably due to the Confucianist child-rearing ideas. These facts are not reflected in other business taxonomies for investment.

To summarize, our method allows \emph{concrete terms} that would not appear in traditional business taxonomies to be displayed and facilitates the \emph{discovery of new terms}. Therefore, the constructed taxonomy has some special advantages in investment activities compared to the static manually designed business classification systems, and can be a meaningful supplementary for the existing business classification systems.

\section{Conclusion} \label{cln}
In this article, we proposed a method to extract concept-level terms with weak and partial supervision and build a taxonomic structure of these terms using greedy hierarchical affinity propagation. The application of this method for business taxonomy construction is novel, for the reason that business texts have different linguistic features to represent ``is-a" relations. 

Our method is fast in both term similarity computing and taxonomy induction. Experiments on the Chinese NEEQ market show that the text-induced business taxonomy has several advantages over the traditional expert-crafted system, such as to display fine-grained concepts and discover trendy business concepts. The method provides a better tool for investment activities and industry research.

Of course, the constructed business taxonomy is not perfect. For instance, the ``Phone gadgets" concept is giant and include too many companies. For this reason, the intra-class similarity is also the lowest for this class. These observations suggest that ``Phone gadgets" can not be a good exemplar for the entire class and the class may be subject to further partition. Additionally, the semantic distances between hypernyms are at different scales: ``Healthcare" and ``Medical service and equipment" are small and related concepts that may be merged. Finally, the other relations between companies within the same set, e. g.~supply chain relations, are not revealed. We will investigate how to improve the taxonomy with these relations in the future.\\
\\
\section*{Appendix} 
Table~\ref{examples} further provides some examples of how label terms generated by our method (GHAP) are different from the NEEQ terms. 

\noindent Contact authors for the full taxonomy structure. 

\newpage

\begin{table*}[t]
\centering
\caption{The NEEQ classification label and label of our method for some companies mapped to the ``Education" concept.} \label{examples}
\resizebox{\textwidth}{!}{
\begin{tabular}{c}
\includegraphics[width=\textwidth]{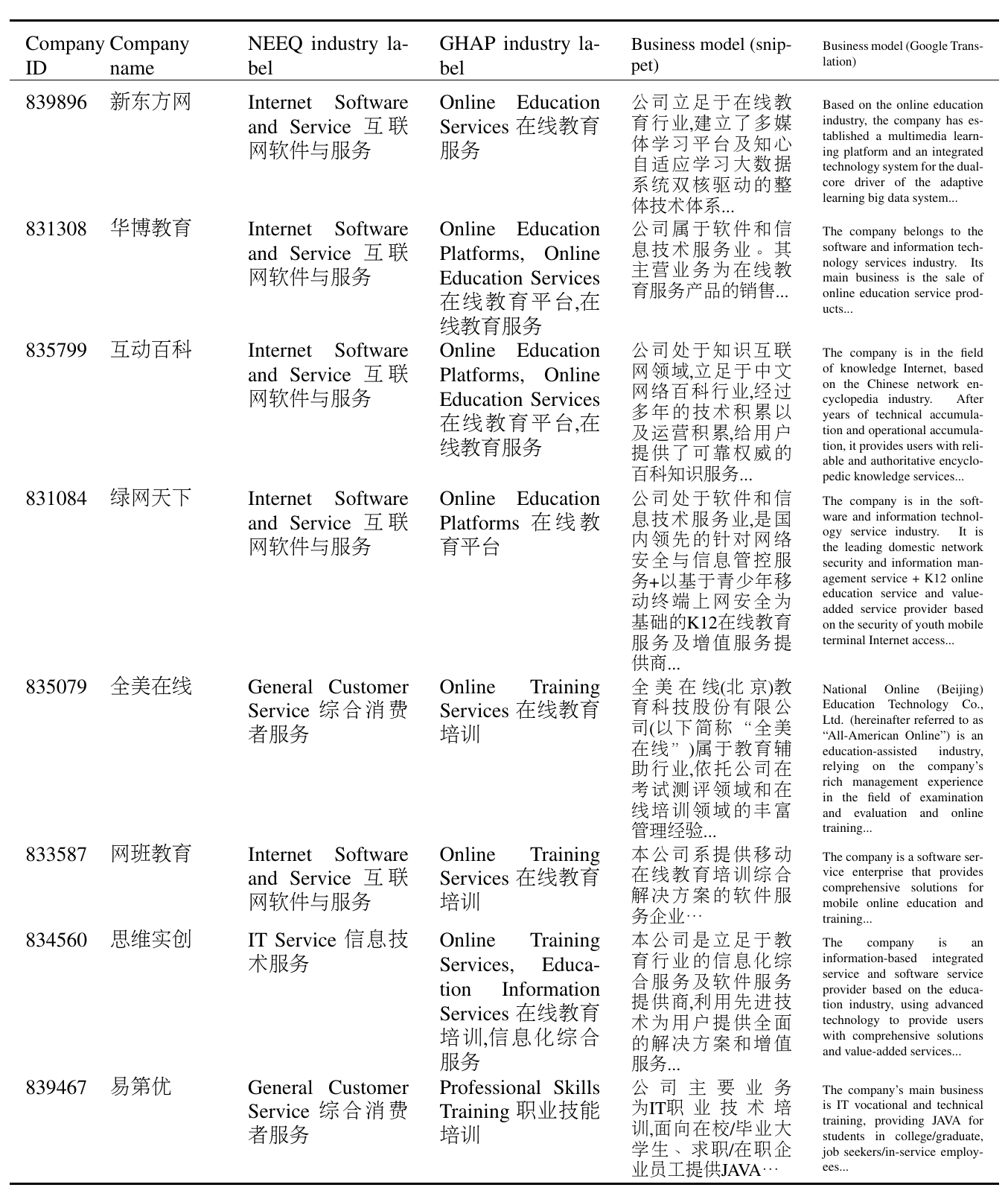}
\end{tabular}}
\end{table*}

\bibliographystyle{named}
\bibliography{frank-btc}

\end{document}